%% file: emnlp2023.tex
\pdfoutput=1

\documentclass[11pt]{article}

\usepackage{EMNLP2023}

\usepackage{times}
\usepackage{latexsym}

\usepackage[T1]{fontenc}

\usepackage[utf8]{inputenc}

\usepackage{microtype}

\usepackage{inconsolata}

\usepackage{booktabs}
\usepackage{makecell}
\usepackage{multirow}
\usepackage{multicol}
\usepackage{pifont}
\newcommand{\cmark}{\ding{51}}%
\usepackage{enumitem}
\usepackage{adjustbox}
\usepackage{graphicx}
\graphicspath{{./figures/}}
\usepackage{xspace}
\usepackage{caption}
\usepackage{subcaption}
\usepackage{enumitem}
\usepackage{amsmath}
\usepackage{amssymb}
\usepackage{cuted}
\usepackage[dvipsnames]{xcolor}

\input{math_command}

\newcommand{\dataname}{DocumentNet}
\newcommand{\modelname}{UniFormer}
\newcommand{\taskname}{VDER}
\definecolor{mygray}{gray}{0.4}

\makeatletter
\DeclareRobustCommand\onedot{\futurelet\@let@token\@onedot}
\def\@onedot{\ifx\@let@token.\else.\null\fi\xspace}

\def\eg{\emph{e.g}\onedot} \def\Eg{\emph{E.g}\onedot}
\def\ie{\emph{i.e}\onedot} 
 
\def\etc{\emph{etc}\onedot}

\makeatother

\newcommand{\cls}{\texttt{[CLS]}\xspace}
\newcommand{\mask}{\texttt{[MASK]}\xspace}

\title{\dataname{}: Bridging the Data Gap in Document Pre-Training}

\author{Lijun Yu$^{\ddagger \diamond}$, Jin Miao$^{\dagger}$, Xiaoyu Sun$^{\dagger}$, Jiayi Chen$^{\nmid}$, Alexander G. Hauptmann$^{\ddagger}$,\\Hanjun Dai$^{\dagger}$, Wei Wei$^{\dagger \diamond}$\\
$^{\ddagger}$Carnegie Mellon University, $^{\dagger}$Google, $^{\nmid}$University of Virginia\\
\texttt{$^{\diamond}$lijun@cmu.edu, wewei@google.com}}

\begin{document}
\maketitle

\begin{abstract}
\input{sections/abstract}
\end{abstract}

\input{tables/dataset}

\section{Introduction}
\input{sections/introduction}
\section{Related Work}
\input{sections/related_work}

\begin{figure*}
    \centering
\begin{tabular}{@{}c@{}c@{}c@{}c@{}}
 \includegraphics[width=0.24\textwidth,clip]{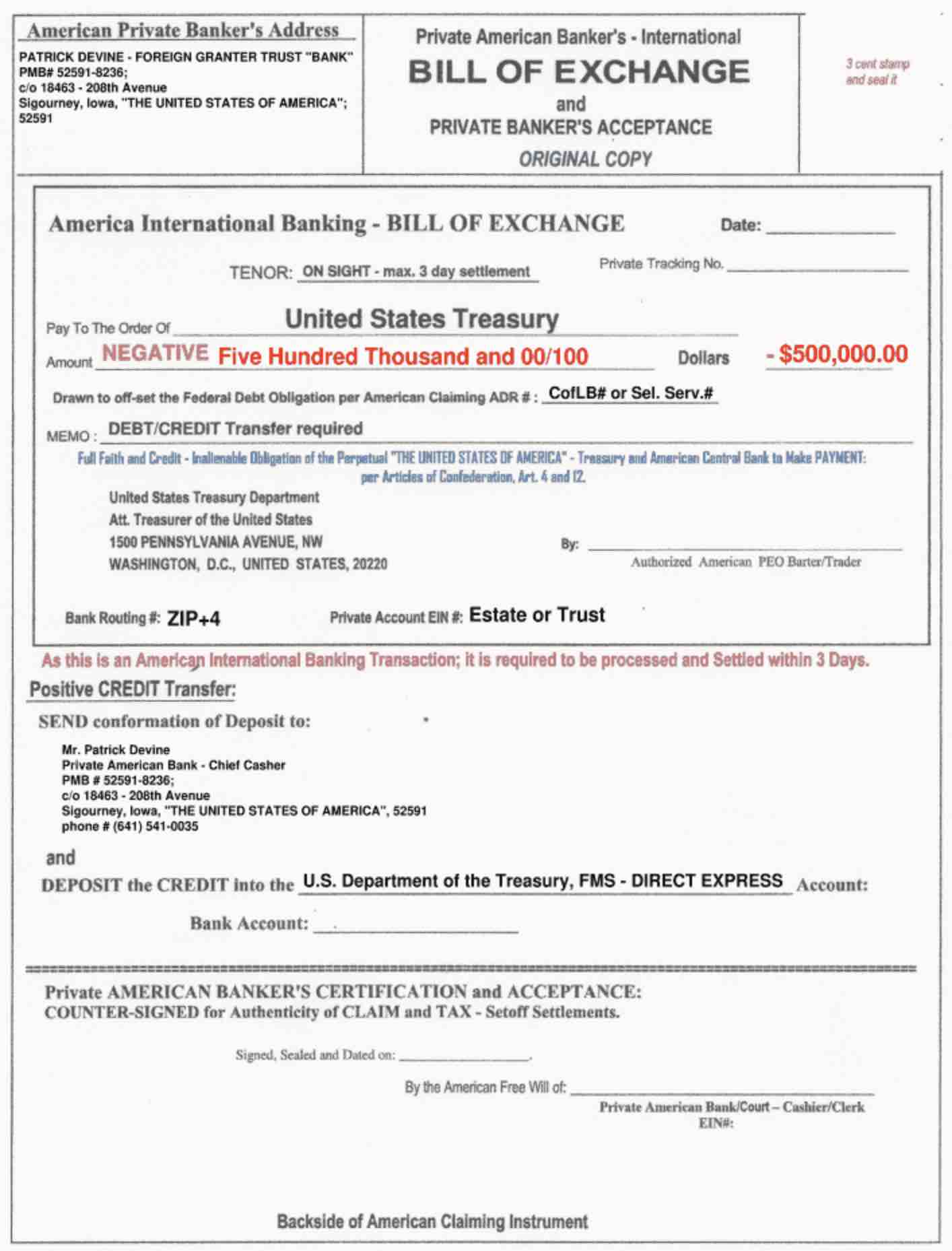} & 
 \includegraphics[width=0.24\textwidth,clip]{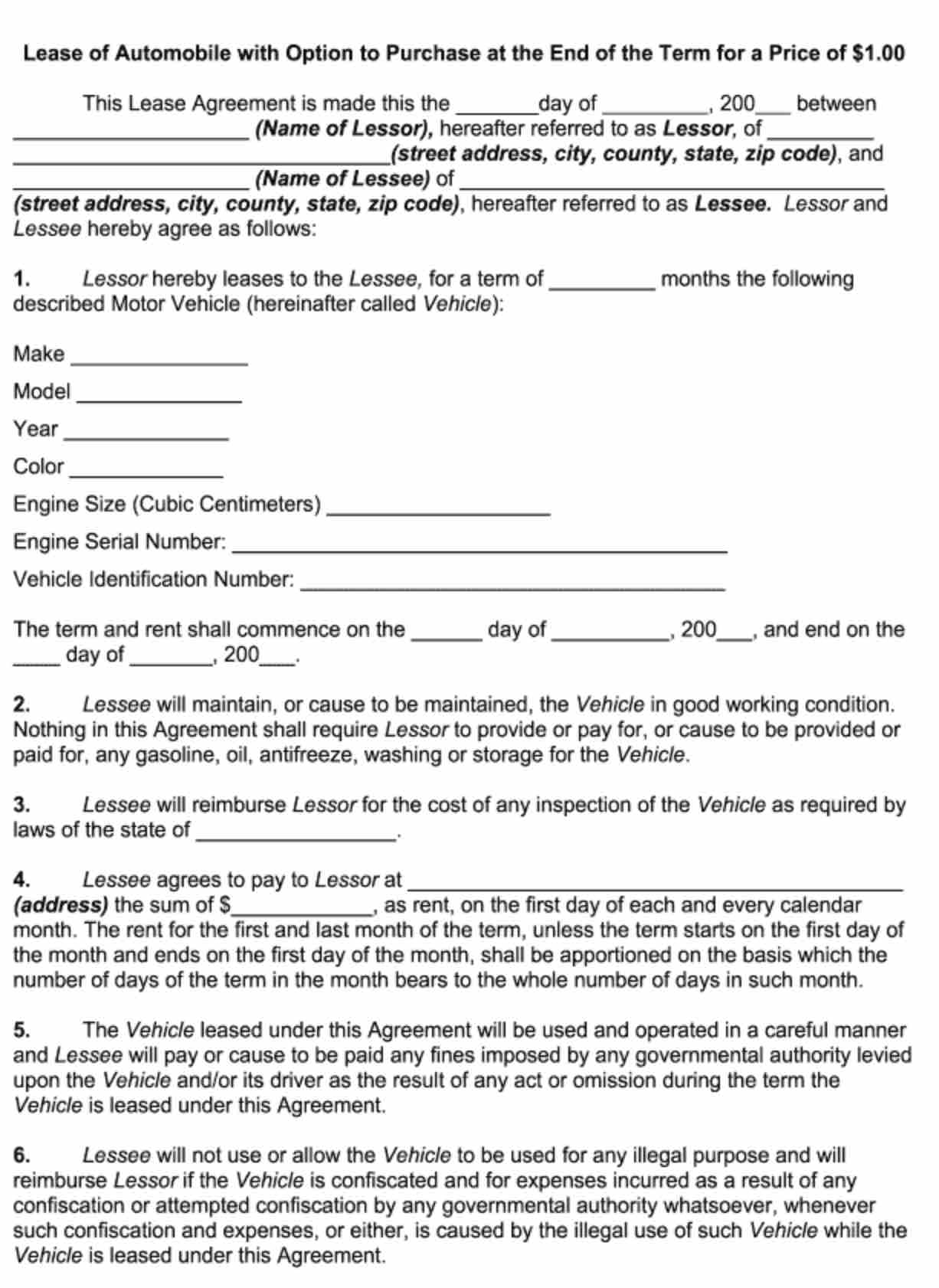} & 
 \includegraphics[width=0.24\textwidth,clip]{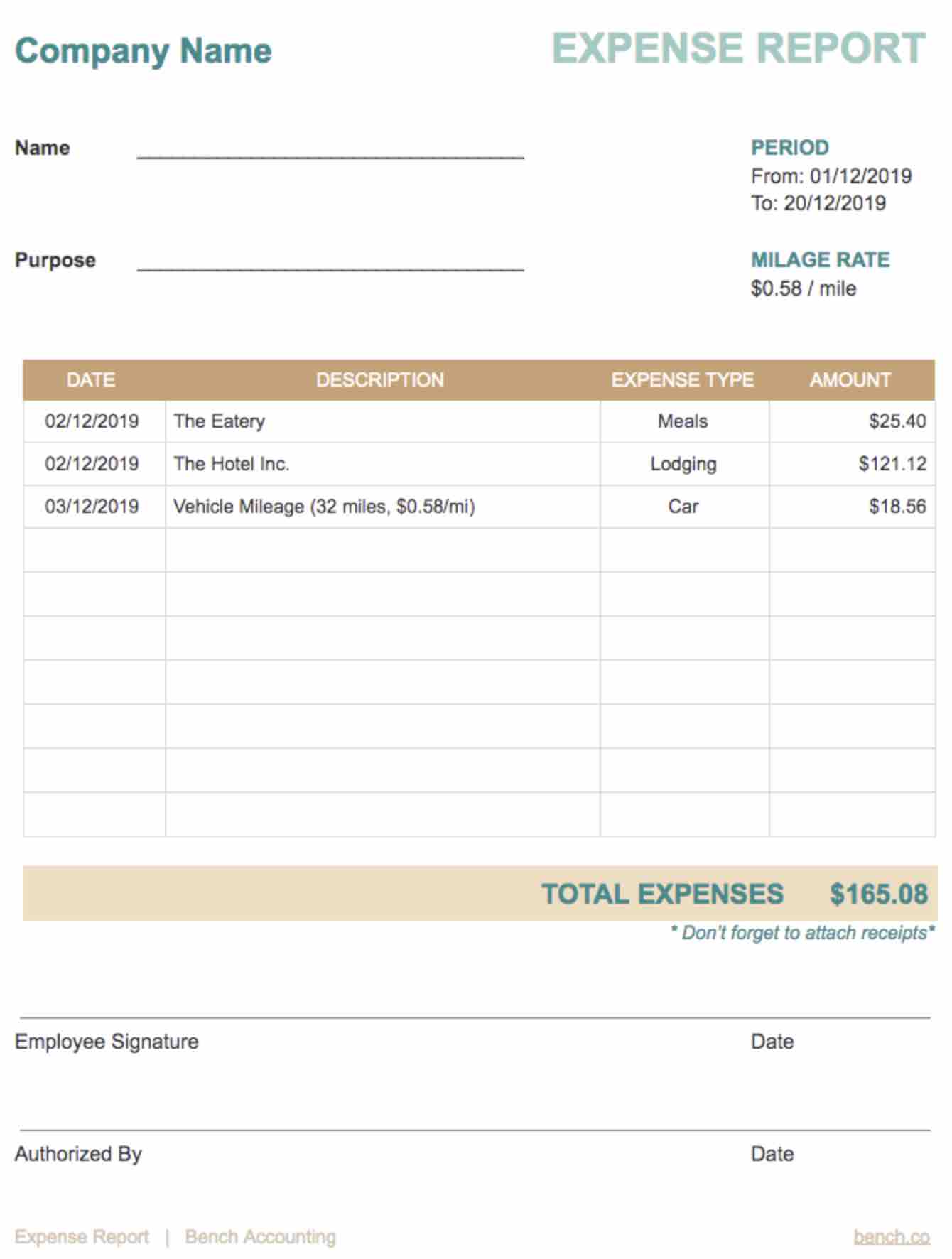} & 
 \includegraphics[width=0.24\textwidth,clip]{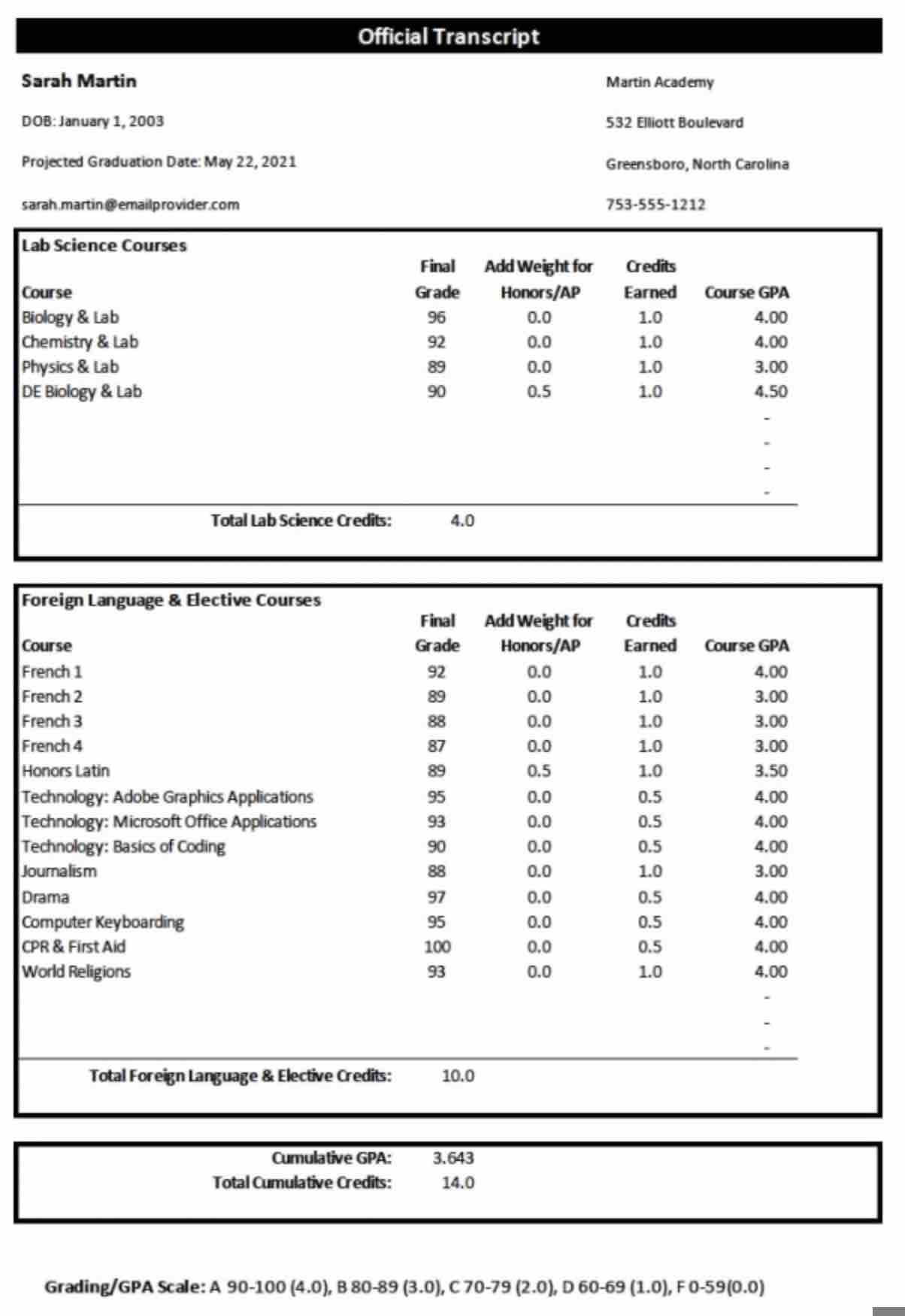} \\
 Financial & Legal & Business & Education
\end{tabular}
    \caption{Exemplar documents of each of the four top-level hierarchies. Images are downloaded via keyword searching using a commercial search engine. All images are for demonstration purposes only and do not contain real transactions or personal information. \label{fig:example}}
\end{figure*}

\section{\dataname{} Dataset}
\input{sections/data_collection}

\section{\modelname{} Model}
\input{sections/model}

\section{Experiments}
\input{sections/experiments}

\section{Conclusions}
\input{sections/conclusions}



\section*{Acknowledgements}
We are grateful for the support from Dale Schuurmans and Evan Huang.
This research was supported in part by the Defence Science and Technology Agency (DSTA).

\bibliography{anthology,custom}
\bibliographystyle{acl_natbib}

\clearpage
\appendix

\input{sections/appendix}

\end{document}

%% file: math_command.tex
\usepackage{amsmath,amsfonts,bm}

\def\1{\bm{1}}

\def\rvc{{\mathbf{c}}}

\def\rve{{\mathbf{e}}}

\def\rvp{{\mathbf{p}}}

\def\rvt{{\mathbf{t}}}

\def\rvv{{\mathbf{v}}}

\def\ervb{{\textnormal{b}}}
\def\ervc{{\textnormal{c}}}

\def\erve{{\textnormal{e}}}

\def\ervp{{\textnormal{p}}}

\def\ervt{{\textnormal{t}}}

\def\ervv{{\textnormal{v}}}

\def\rmD{{\mathbf{D}}}

\DeclareMathAlphabet{\mathsfit}{\encodingdefault}{\sfdefault}{m}{sl}
\SetMathAlphabet{\mathsfit}{bold}{\encodingdefault}{\sfdefault}{bx}{n}

\def\gM{{\mathcal{M}}}

\def\gO{{\mathcal{O}}}

\def\gT{{\mathcal{T}}}

\def\sR{{\mathbb{R}}}

\newcommand{\E}{\mathbb{E}}
\newcommand{\Ls}{\mathcal{L}}



%% file: sections/abstract.tex
Document understanding tasks, in particular, Visually-rich Document Entity Retrieval (\taskname{}), have gained significant attention in recent years thanks to their broad applications in enterprise AI.
However, publicly available data have been scarce for these tasks due to strict privacy constraints and high annotation costs.
To make things worse, the non-overlapping entity spaces from different datasets hinder the knowledge transfer between document types.
In this paper, we propose a method to collect massive-scale and weakly labeled data from the web to benefit the training of \taskname{} models. 
The collected dataset, named \dataname{}, does not depend on specific document types or entity sets, making it universally applicable to all \taskname{} tasks.
The current \dataname{} consists of 30M documents spanning nearly 400 document types organized in a four-level ontology.
Experiments on a set of broadly adopted VDER tasks show significant improvements when \dataname{} is incorporated into the pre-training for both classic and few-shot learning settings.
With the recent emergence of large language models (LLMs), \dataname{} provides a large data source to extend their multimodal capabilities for VDER.

%% file: tables/dataset.tex
\begin{table*}[tp]
\centering
\begin{adjustbox}{max width=\textwidth}
\begin{tabular}{lccccc}
\toprule
Dataset & \makecell{\#Samples$\uparrow$} & \makecell{Ontology} & \makecell{Diverse\\Domains} & \makecell{High-quality\\OCR} & \makecell{Annotation} \\ \midrule
FUNSD~\cite{jaume2019funsd}  & 199 &       &      & & E=3  \\
Kleister-NDA~\cite{stanislawek2021kleister}   &  540 &       &      & \cmark & E=4 \\ 
VRDU-Ad-buy~\cite{wang2022benchmark}   & 641 &       &      & \cmark & E=14 \\
SROIE~\cite{huang2019icdar2019}    & 973  &  &      & & E=4 \\
CORD~\cite{park2019cord}  &  1K &       &      & & E=30 \\
DeepForm~\cite{borchmann2021due}   &  1.1K &       &      & \cmark & E=5  \\ 
VRDU-Registration~\cite{wang2022benchmark}   &  1.9K &       &      & \cmark & E=6 \\ 
Kleister-Charity~\cite{stanislawek2021kleister}   &  2.7K  &       &      & \cmark & E=8 \\
DocVQA~\cite{mathew2021docvqa} & 12.8K & & & \cmark & Q \\ \midrule
CC-PDF~\cite{powalski2021going} & 350K & & \cmark & & \\
PubLayNet~\cite{zhong2019publaynet} & 358K & & \cmark&  & B=5 \\
RVL-CDIP~\cite{lewis2006building}   &  400K  &    & \cmark  &   & C=16 \\ 
UCSF-IDL~\cite{powalski2021going} & 480K & & \cmark & & \\
IIT-CDIP~\cite{lewis2006building}   &  11.4M  &  & \cmark     &  \\
\midrule
\color{mygray}ImageNet~\cite{deng2009imagenet} & \color{mygray}1.3M images & \color{mygray}\cmark & \color{mygray}- & \color{mygray}- & \color{mygray} C=1K \\
\color{mygray}ActivityNet~\cite{caba2015activitynet} & \color{mygray}20K videos & \color{mygray}\cmark & \color{mygray}- & \color{mygray}- & \color{mygray} C=200 \\ \midrule
\emph{\dataname-v1 (ours)}  & {9.9M}      &  \cmark     &   \cmark   & \cmark & \textbf{C=398, E=6} \\
\emph{\dataname-v2 (ours)}  & \textbf{30M}      &  \cmark     &   \cmark   & \cmark & \textbf{C=398, E=6} \\ \bottomrule
\end{tabular}
\end{adjustbox}
\vspace{-2mm}
\caption{Comparison between the proposed \dataname{} dataset and existing document understanding datasets. Datasets from other areas also built with ontology are listed in \textcolor{mygray}{gray}. Annotation includes class label (C), bounding box (B), entity (E), and question (Q), where the value refers to the number of classes.}
\label{tab:datasets}
\vspace{-2mm}
\end{table*}


%% file: sections/introduction.tex
Document understanding is one of the most error-prone and tedious tasks many people have to handle every day.
Advancements in machine learning techniques have made it possible to automate such tasks.
In a typical Visually-rich Document Entity Retrieval (VDER) task, pieces of information are retrieved from the document based on a set of pre-defined entity types, known as the \textit{schema}.
For example, ``amount'', ``date'', and ``item name'' are major parts of an invoice schema.

The current setup of VDER tasks presents several unique challenges for acquiring sufficient training data.
First, the availability of raw document images is greatly limited due to privacy constraints.
Real-world documents, such as a driver's license or a bank statement, often contain personally identifiable information and are subject to access controls.
Second, detailed annotation is costly and typically requires intensive training for experienced human annotators.
\Eg, it takes deep domain knowledge to correctly label different fields in complex tax forms.
Finally, knowledge sharing between various types of documents is constrained by inconsistent label spaces and contextual logic.
For example, the entity sets (\ie, schema) could be mutually exclusive, or the same entity type could take different semantic meanings in different contexts.

A number of models have been proposed for VDER tasks with various success~\cite{huang2022layoutlmv3,lee2022formnet,appalaraju2021docformer,gu2021unidoc}.
To tackle the aforementioned challenges, most prior works initialize from a language model followed by BERT-style~\cite{devlin2019bert} pre-training on document datasets with additional layout and visual features.
However, even the largest dataset currently in use, \ie IIT-CDIP~\cite{lewis2006building} dataset, has a limited size and only reflects a subset of document types.

In this paper, we introduce the method of building the \dataname{} dataset, which enables massive-scale pre-training for VDER modeling.
\dataname{} is collected over the Internet using a pre-defined ontology, which spans hundreds of document types with a four-level hierarchy.
Experiments demonstrated that \dataname{} is the key to advancing the performance on the commonly used FUNSD~\cite{jaume2019funsd}, CORD~\cite{park2019cord}, and RVL-CDIP~\cite{lewis2006building} benchmarks in both classic and few-shot setups.
More recently, LLMs \cite{openai2023gpt4, googlepalm2} have shown great potential for VDER tasks given their reasoning capabilities.
\dataname{} provides massive-scale multimodal data to boost the performance of LLMs for document understanding.

%% file: sections/related_work.tex

Tab.~\ref{tab:datasets} provides an overview of relevant document datasets, with more details in App.~\ref{app:related_data}.

\vspace{-2mm}
\paragraph{Single-domain document datasets.}
Many small document datasets with entity-span annotations have been used for tasks such as entity extraction.
They contain less than 100k pages from a single domain.
Newer datasets come with high-quality OCR annotation thanks to the advantage of relevant tools, while older ones, such as FUNSD~\cite{jaume2019funsd}, often contain OCR errors.
These datasets do not contain sufficient samples for the pre-training of a large model.

\vspace{-2mm}
\paragraph{Large document datasets.}
A few larger datasets contain over 100k pages from different domains. 
However, they usually do not contain OCR annotations or entity-level labels.
IIT-CDIP~\cite{lewis2006building} has been the largest dataset commonly used for pre-training of document understanding models.
Although these datasets are large, their image quality and annotation completeness are often unsatisfactory.
To complement them, we collect high-quality document images from the Internet to build the \dataname{} datasets with rich OCR and entity annotations, and demonstrate their effectiveness in document model pre-training.

\vspace{-2mm}
\paragraph{Ontology-based datasets.}
Large labeled datasets are usually collected following an ontology.
ImageNet~\cite{deng2009imagenet} for image recognition is built upon the synsets of WordNet~\cite{miller1998wordnet}.
ActivityNet~\cite{caba2015activitynet} for activity recognition adopts an activity taxonomy with four levels.
To the best of our knowledge, \dataname{} is the first large-scale document dataset built upon a well-defined ontology.

\vspace{-2mm}
\paragraph{Pretrained document models.}
A variety of pretrained document models have emerged, including LayoutLM~\cite{xu2020layoutlm},  UDoc~\cite{gu2021unidoc}, LayoutLMv2~\cite{xu2021layoutlmv2}, TILT~\cite{powalski2021going}, BROS~\cite{hong2022bros}, DocFormer~\cite{appalaraju2021docformer}, SelfDoc~\cite{li2021selfdoc}, LayoutLMv3~\cite{huang2022layoutlmv3}, \etc.
App.~\ref{app:related_model} provides detailed comparisons of their designs.




%% file: sections/data_collection.tex
\begin{figure}[tp]
    \centering
    \includegraphics[width=\linewidth, trim={0.2cm 0 0.2cm 0}, clip]{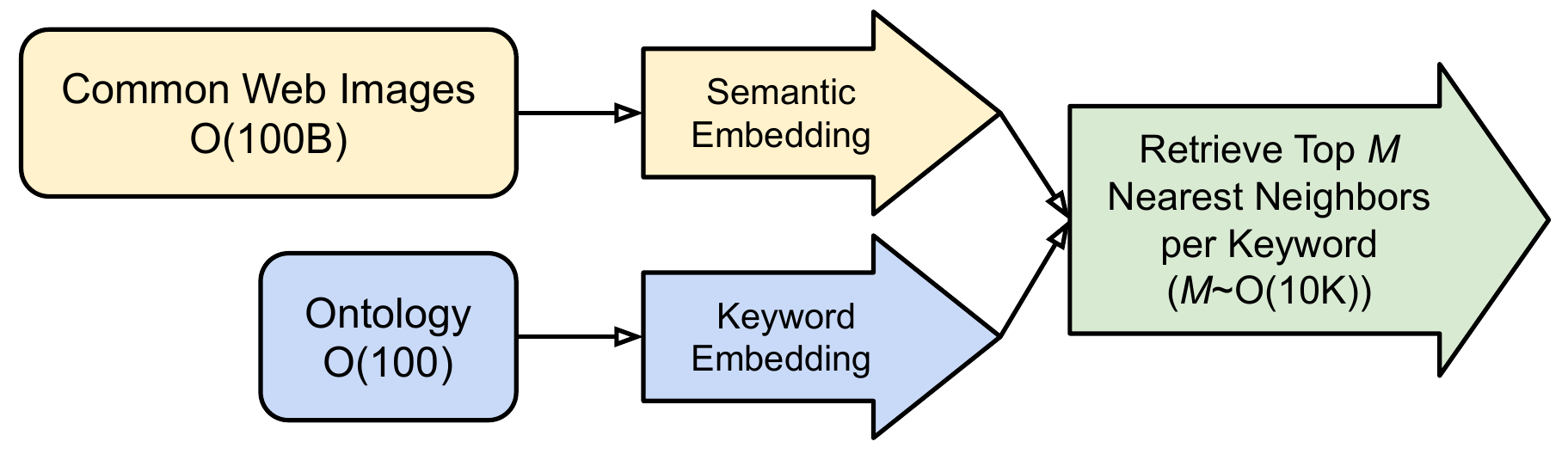}
    \vspace{-2mm}
    \caption{Data Collection Pipeline.}
    \label{fig:data_pipeline}
    \vspace{-2mm}
\end{figure}

Blindly crawling the Web for images may seem easy, but it is not a practical solution since most images on the Web are not relevant to document types.
We need a scalable pipeline to only select the concerned images.
Broadly, this is achievable via a nearest-neighbor search of relevant keywords in a text-image joint embedding space.
First, we design a set of query keywords in English, \ie, the document ontology, and encode them into the embedding space of general Web images. 
Further, a nearest-neighbor algorithm retrieves the top-K semantically closest images to each query keyword. 
Finally, a deduplication step consolidates all retrieved images across all query keywords. 
Fig.~\ref{fig:example} illustrates several exemplar documents retrieved using our provided keywords. 

\vspace{-2mm}
\paragraph{Ontology creation.}
Each text string in the ontology list serves as a seed to retrieve the most relevant images from the general Web image pool. An ideal ontology list should therefore cover a broad spectrum of query keywords across and within the concerned downstream application domains. Although algorithmic or generative approaches may exist, in this paper, we manually curated about 400 document-related query keywords that cover domains of finance, business, personal affairs, legal affairs, tax, education, etc.
The full ontology hierarchy and keyword list are provided in App.~\ref{app:ontology}.

\begin{figure}[tp]
    \centering
    \includegraphics[width=\linewidth, trim={0.2cm 0 0.2cm 0}, clip]{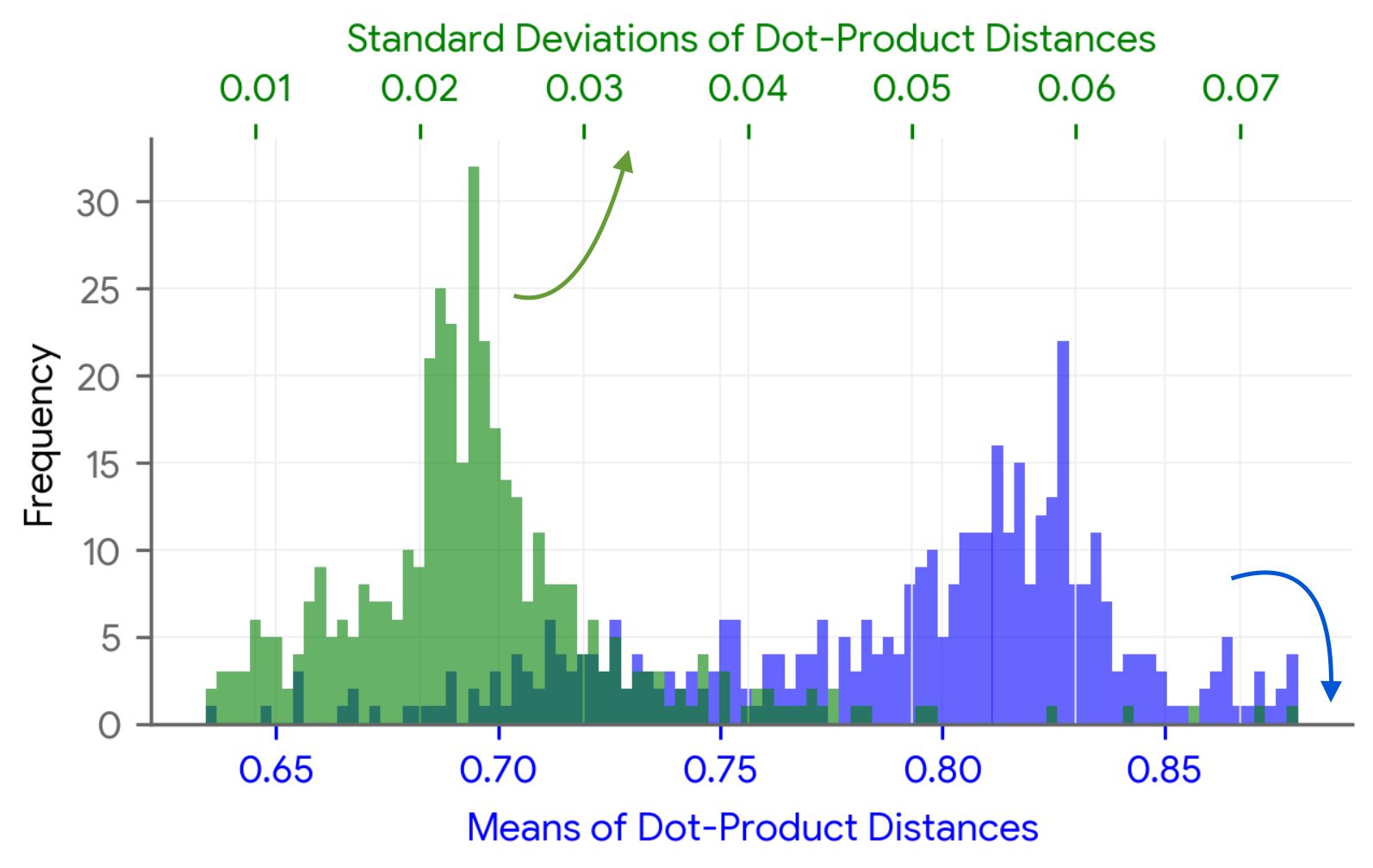}
    \vspace{-4mm}
    \caption{\textcolor{blue}{Mean} and \textcolor{ForestGreen}{standard deviation} of the dot-product distance between the retrieved 30M document images and each query keyword. A distance of $1.0$ indicates the closest semantic relevance.}
    \label{fig:dataset_distance_histo}
    \vspace{-2mm}
\end{figure}

\begin{figure*}[tp]
    \centering
    \includegraphics[width=\linewidth,trim={0 0 1.2cm 0},clip]{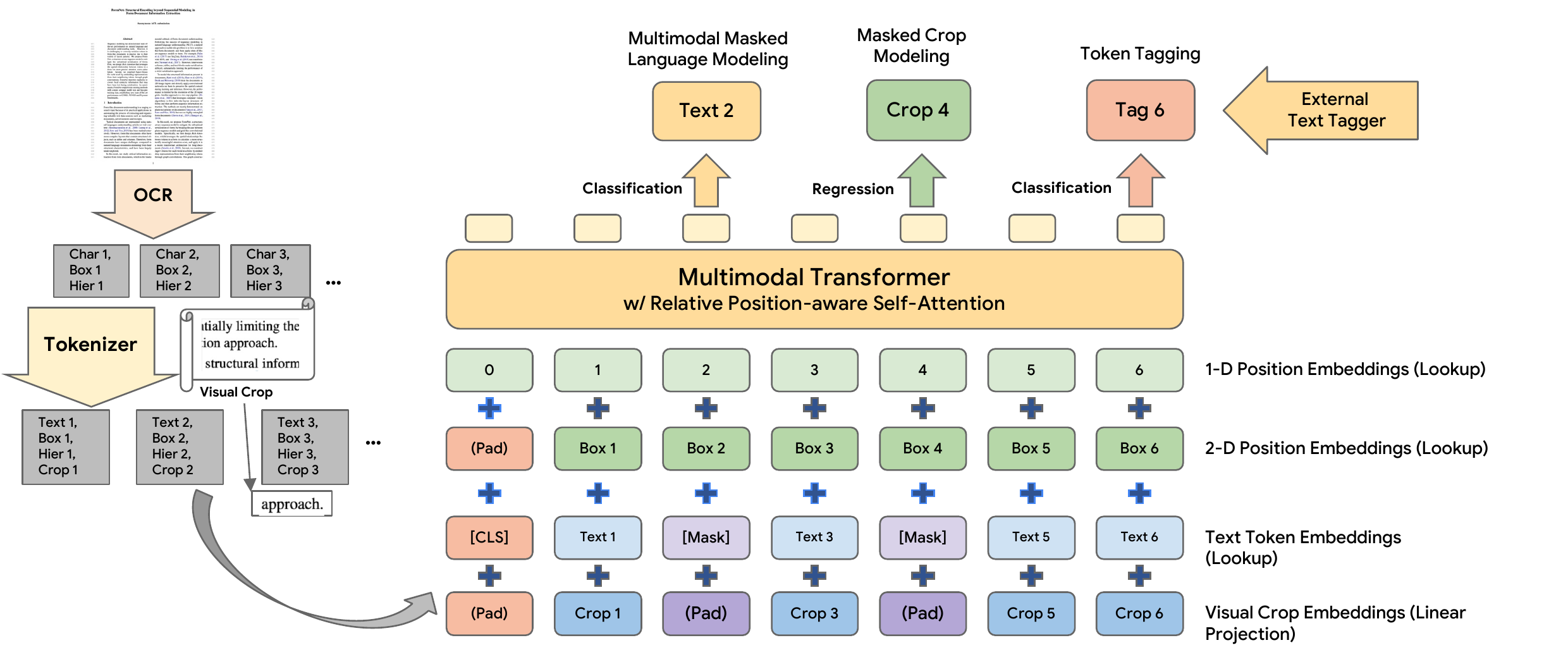}
    \vspace{-2mm}
    \caption{\modelname{} pre-training pipeline. The multimodal tokenization process (left) outputs tokens with aligned image crops. The \modelname{} model (right) learns a unified token representation with three objectives (top).}
    \label{fig:pretrain_arch}
    \vspace{-2mm}
\end{figure*}

\vspace{-2mm}
\paragraph{Image retrieval from ontology.}
To retrieve only the most relevant document images out of the hundreds of billions of general Web images, we leverage a highly efficient nearest neighbor pipeline by defining the similarity metric as the dot product between the semantic feature vectors of the image and each of the target query keywords. Here we refer to Graph-RISE~\cite{juan2020imageembed} for the semantic image embedding, and all query keywords are encoded into the same feature space as the images.
Empirically, we pick the top 10k nearest neighbors in English for each query keyword. Note that the same image might be retrieved via multiple semantically similar keywords, so a de-duplication step is needed afterward. We summarize the main pipeline steps in Fig.~\ref{fig:data_pipeline}.
Fig.~\ref{fig:dataset_distance_histo} shows statistical insights of the retrieved 30M document images with the mean and standard deviation histogram over each of the query keywords.
The majority of the retrieved images are with mean distance values greater than $0.8$ and standard deviations no more than $0.03$, indicating high relevance to the document ontology.

\input{tables/pretrain}
\vspace{-2mm}
\paragraph{OCR and annotation.}
\label{sec:ocr}
The retrieved images are fed into an OCR engine to generate a text sequence in reading order.
We apply a text tagging model to weakly annotate the text segments of each sequence into 6 classes, including \textit{email addresses, mail addresses, prices, dates, phone numbers,} and \textit{person names}. 
Albeit noisy, these classification labels provide additional supervision for pre-training.

\vspace{-2mm}
\paragraph{Post-processing and open-source tools.}
We adopt some heuristic-based filtering to improve sample quality.
For example, we remove samples where the overall OCR result is poor due to blurry or noisy images.
Some proprietary tools are used for scalable processing during the construction of \dataname{}, but open-source alternatives are readily available.
\Eg, CLIP~\cite{radford2021learning} for text-image embedding, Google ScaNN~\cite{avq_2020} for scalable nearest-neighbor search, Google Cloud OCR~(\url{https://cloud.google.com/vision/docs/ocr}), and Google Cloud NLP~(\url{https://cloud.google.com/natural-language/docs/reference/rest/v1/Entity#type}) for text tagging.

With all of the above steps, we have obtained a dataset of high-quality document images that are closely relevant to our query ontology. 
This dataset contains multiple modalities, including the image pixels, the OCR characters, the layout coordinates, and the segment tags.

%% file: tables/pretrain.tex
\begin{table}[tp]
\centering
\begin{adjustbox}{max width=\linewidth}
\begin{tabular}{lll}
\toprule
 & Task                                       & Target Modality    \\
\midrule
MMLM & \makecell[l]{Multimodal Masked\\\ \ Language Modeling} & OCR characters     \\
MCM & Masked Crop Modeling            & Image pixels    \\
TT & Token Tagging                 & Segment tags \\
\bottomrule
\end{tabular}
\end{adjustbox}
\vspace{-2mm}
\caption{\modelname{} pre-training objectives and corresponding target modalities.}
\label{tab:pretrain}
\vspace{-2mm}
\end{table}

%% file: sections/model.tex
To take advantage of all the modalities available in \dataname{},
we build a lightweight transformer model named \modelname{} for document pre-training.
Table \ref{tab:pretrain} lists the pre-training objectives and corresponding target modalities.

\modelname{} is built upon the BERT~\cite{devlin2019bert} architecture similar to LayoutLM~\cite{xu2020layoutlm} and LayoutLMv2~\cite{xu2021layoutlmv2}.
Figure \ref{fig:pretrain_arch} illustrates the pre-training pipeline.
We highlight the new designs for multimodal pretraining here and defer more details into App. \ref{app:model}.

\input{tables/ablation}
\vspace{-2mm}
\paragraph{Multimodal tokenization and embedding.}
With a pre-defined text tokenizer, \eg WordPiece~\cite{wu2016google}, we first tokenize the OCR characters into a sequence of text tokens $\rvc$.
For each token $\ervc_i$, we obtain its bounding box $\ervb_i = (x_0, y_0, x_1, y_1)_i$ by taking the union of the bounding boxes of its characters.
We enlarge the bounding box by a context ratio $r$ on each side and obtain the corresponding visual image crop $\ervv_i$ for each token from the raw image.
To model visual information, we add a crop embedding by linearly projecting the flattened pixels in the image crop, following ViT~\cite{dosovitskiy2020image}.

\vspace{-2mm}
\paragraph{Masked crop modeling.}
In addition to predicting the text token in the MMLM objective,
A \modelname{} parameterized by $\theta$ also predicts the visual modality by reconstructing the image crops for the masked tokens, in a way similar to MAE~\cite{he2022masked}.
It is formulated as a regression problem with a linear layer outputing flattened pixels and the objective is
\vspace{-2mm}
\begin{equation}
    \Ls_\mathit{MCM} = \mathop{\E}_{\text{data}} \Big[ \sum_{\ervv_i \in \gM} \left\|{f_\theta(\overline{\rvc}, \overline{\rvv}, \rho)}_i - \ervv_i\right\|_2^2 \Big]
\end{equation}
where $\overline{\rvc}$ and $\overline{\rvv}$ denote the masked tokens and crops according to mask $\gM$.
$\rho$ is the position and layout embeddings.

\vspace{-2mm}
\paragraph{Token tagging.}
With fully unmasked sequences, \modelname{} is pre-trained to predict the token tags $\rvt$ with a separate head.
Since each token may have multiple tags, it is formulated as a multi-label classification problem with binary cross-entropy losses.

%% file: tables/ablation.tex
\begin{table*}[tp]
\centering
\begin{adjustbox}{max width=\textwidth}
\begin{tabular}{llllccc}
\toprule
\multirow{2}{*}{Model} & \multirow{2}{*}{Inputs}  & \makecell[l]{Pre-training\\Data} & \makecell[l]{Pre-training\\Objectives} & {\makecell{FUNSD\\Entity F1$\uparrow$}}  & {\makecell{CORD\\Entity F1$\uparrow$}} & {\makecell{RVL-CDIP\\Accuracy$\uparrow$}}  \\ \midrule

BERT & T  & -  & MLM & 60.26 & 89.68 & 89.81 \\

LayoutLM & T + L & IIT-CDIP & MVLM & 78.66 & 94.72 & 91.78 \\ 

\makecell[l]{\emph{\modelname{}}} & T + L + C & IIT-CDIP                & MMLM        & 80.63 & 95.17 & 93.47   \\ \midrule
\multirow{3}{*}{\makecell[l]{\emph{\modelname{}}}} & \multirow{3}{*}{\makecell{T + L + C}} &  \multirow{3}{*}{\makecell[l]{IIT-CDIP\\\ \ + \emph{\dataname-v1}}}   & MMLM   & 82.61 & 95.91 & 94.86   \\
& &          & MMLM + MCM & 83.45 & 96.08 & 95.15   \\ 
 &  &   &    MMLM + MCM + TT    & \textbf{84.18} & \textbf{96.45} & \textbf{95.34}   \\
\bottomrule
\end{tabular}
\end{adjustbox}
\caption{Ablation studies on three document understanding benchmarks regarding pretraining datasets, pretraining objectives, and model architectures. Input modalities include text (T), layout (L), and crop (C).}
\label{tab:ablation}
\vspace{-2mm}
\end{table*}

%% file: sections/experiments.tex
\input{tables/sota}
\input{tables/few-shot}

We pre-train \modelname{} on \dataname{} and evaluate on two settings:
(1) the classic VDER setting with the full split of train and test;
(2) the few-shot VDER setting where we have meta-train and meta-test task sets with each task containing a set of samples that satisfies the $N$-way $K$-shot setting. 

\vspace{-2mm}
\subsection{Pre-Training}
We initialize our \modelname{} with BERT weights using the uncased vocabulary.
The models are pre-trained using the Adam optimizer~\cite{kingma2014adam}.
We adopt a cosine learning rate schedule with linear warmup during the first 2\% steps and a peak learning rate of $10^{-4}$.
We use 20\% of the samples for the token tagging pre-training task.
The models are trained for 500K steps with a batch size of 2048 on 128 TPUv3 devices.



\subsection{Classic VDER Setting}
We evaluate the performance of pre-trained \modelname{} models on three commonly used benchmarks: entity extraction on FUNSD and CORD, and document classification on RVL-CDIP.
Detailed setups are provided in App. \ref{app:classic}.

\vspace{-2mm}
\paragraph{Implementation details.}
For entity extraction on FUNSD and CORD, we add a \emph{Simple} multi-class classification head on top of all text tokens to perform BIO tagging.
We fine-tune with a peak learning rate of $5\times10^{-5}$, following a schedule of linear warm-up in the first 10\% steps and then linear decay. Dropout with 0.1 probability is applied in the head layers. \modelname{} is fine-tuned for 1000 steps with a batch size of 32 on FUNSD and 256 on CORD.
For document classification on RVL-CDIP, we add a multi-class classification head on top of the [CLS] token. We fine-tune with a constant learning rate of $10^{-5}$ for 15000 steps with a batch size of 2048.

\vspace{-2mm}
\paragraph{Ablation Studies.}
Table \ref{tab:ablation} lists the ablation results for pre-training data, pre-training objectives, and model design.
Compared to LayoutLM, our unified embedding of the visual modality and MMLM pre-training results in a much stronger baseline.
Adding our \dataname{} into the pre-training leads to a significant performance boost across all three tasks.
Further incorporating MCM and TT pre-training objectives to fully leverage \dataname{} yields consistent improvements, where the entity extraction tasks benefit more from TT and the document classification task gains more from MCM.

\vspace{-2mm}
\paragraph{Comparisons with existing methods.}
We compare the performance on the three benchmarks with existing approaches at the base model scale in Table \ref{tab:sota}.
As shown, most prior methods use stronger language or image initialization compared to our lightweight \modelname{}, but all of them are only pre-trained on datasets no larger than IIT-CDIP.
Although \modelname{} is only using 115M parameters and BERT initialization,
it outperforms all baseline approaches after pre-training on our \dataname{} dataset, with FUNSD entity F1 84.18, CORD entity F1 96.45, and RVL-CDIP accuracy 95.34.

\subsection{Few-shot VDER Setting}
We evaluate the performance of pre-trained \modelname{} models on N-way K-shot meta-learning settings with the CORD dataset.
Detailed task setups are introduced in App. \ref{app:few-shot}.

\vspace{-2mm}
\paragraph{Implementation details.}
In addition to the \emph{Simple} prediction head used in the classic setting, we also adopt a two-level \emph{Hierarchical} prediction head.
At the first level, it does a binary classification of the O-tag to identify background tokens.
Non-background tokens are further classified by the second level.
Hierarchical prediction helps reduce the label imbalance problem where the majority of the tokens are labeled as background.
After eliminating a few entities that do not appear frequently enough, we use 18 entities for meta-train and 5 entities for meta-test, for a total of 23 entities.
We fine-tune for 15 steps with a constant learning rate of 0.02.

\vspace{-2mm}
\paragraph{Results.}
As shown in Tab \ref{tab:few-shot}, adding the \dataname{} data significantly boosts the performance of our models across all few-shot learning settings.
In particular, the 30M \dataname{}-v2 variant yields a much larger improvement than the 9.9M \dataname{}-v1.
The amount of data and the diversity in terms of the collected document type played a significant role in the performance improvements. Performance improvements are universal across each of the metrics, with recall improvements more significant than precision.

%% file: tables/sota.tex
\begin{table*}[t]
\centering
\begin{adjustbox}{max width=\textwidth}
\begin{tabular}{@{}l@{\hspace{8pt}}l@{\hspace{8pt}}c@{\hspace{8pt}}l@{\hspace{8pt}}c@{\hspace{8pt}}c@{\hspace{8pt}}c@{}}
\toprule
Model & Initialization & \makecell{Total\\Parameters} & \makecell{Pretrain\\Data Source}           & \makecell{FUNSD\\Entity F1$\uparrow$}  & \makecell{CORD\\Entity F1$\uparrow$} & \makecell{RVL-CDIP\\Accuracy$\uparrow$}  \\ \midrule


\multirow{2}{*}{\makecell[l]{LayoutLM}} & BERT & 113M & IIT-CDIP  & 78.66 & 94.72 & 91.78 \\
 & \makecell[l]{BERT + ResNet-101} & 160M &  IIT-CDIP  & 79.27 & - & 94.42 \\
\makecell[l]{UDoc} & \makecell[l]{BERT + ResNet-50} & 272M & IIT-CDIP  & - & - & 95.05 \\
\makecell[l]{LayoutLMv2} & \makecell[l]{UniLM + ResNeXt-101} & 200M & IIT-CDIP  & 82.76 & 94.95 & 95.25 \\
\makecell[l]{TILT} & \makecell[l]{T5 + U-Net} & 230M & \makecell[l]{RVL-CDIP + \\\ \ \ UCSF-IDL + \\\ \ \ CC-PDF}  & - & 95.11 & 95.25 \\
\makecell[l]{BROS} & BERT & 110M & IIT-CDIP  & 83.05 & 95.73 & - \\
\makecell[l]{DocFormer} & \makecell[l]{LayoutLM + ResNet-50} & 183M & IIT-CDIP  & 83.34 & 96.33 & 96.17 \\
\makecell[l]{SelfDoc} & \makecell[l]{BERT + ResNeXt-101} & 137M & RVL-CDIP  & 83.36 & - & 92.81 \\
\makecell[l]{LayoutLMv3}$^*$ & \makecell[l]{RoBERTa} & 126M & IIT-CDIP  & - & 96.11 & 95.00 \\
\midrule
\makecell[l]{\emph{\modelname{}}} & BERT & 115M & \makecell[l]{IIT-CDIP +\\ \ \ \ \emph{\dataname-v1}}          & \textbf{84.18} & \textbf{96.45} & 95.34   \\ 
\bottomrule
\end{tabular}
\end{adjustbox}
\vspace{-2mm}
\caption{Comparison with existing document pretraining approaches on three document understanding benchmarks. Models at the base scale are listed for fair comparisons, while state-of-the-art results are obtained by models at larger scales. $^*$ denotes a variant that does not use its proprietary tokenizer in pre-training.}
\label{tab:sota}
\end{table*}

%% file: tables/few-shot.tex
\begin{table*}[tp]
\centering
\begin{adjustbox}{max width=\textwidth}
\begin{tabular}{ll | ccc |ccc |ccc}

\toprule
 Datasets & \makecell[r]{Setting\\Prediction Head}  &   \multicolumn{3}{c}{\makecell{4-way 2-shot\\Simple}}    & \multicolumn{3}{c}{\makecell{4-way 2-shot\\Hierarchical}} & \multicolumn{3}{c}{\makecell{4-way 4-shot\\Hierarchical}} \\
\midrule
& & F1 & Prec. & Recall & F1 & Prec. & Recall & F1 & Prec. & Recall \\
\multicolumn{2}{l|}{IIT-CDIP} & 0.099& 0.253& 0.062 &0.108& 0.103& 0.114 & 0.115 & 0.110 & 0.123 \\
\multicolumn{2}{l|}{IIT-CDIP + \emph{\dataname-v1}} & 0.102& 0.217 & 0.067  &0.121 & 0.114 &  0.132  &0.129& 0.125 & 0.134\\
\multicolumn{2}{l|}{IIT-CDIP + \emph{\dataname-v2}} &  \textbf{0.133}&\textbf{0.263}&\textbf{0.090} & \textbf{0.147}&\textbf{0.137}&\textbf{0.160} & \textbf{0.157}&\textbf{0.155}&\textbf{0.160}\\
\bottomrule
\end{tabular}
\end{adjustbox}
\vspace{-2mm}
\caption{Performance comparisons on the few-shot VDER settings with the CORD dataset.
}
\vspace{-3mm}
\label{tab:few-shot}
\end{table*}

%% file: sections/conclusions.tex
In this paper, we proposed a method to use massive and noisy web data to benefit the training of VDER models. Our approach has the benefits of providing a large amount of document data with little cost compared to usual data collection processes in the VDER domain. Our experiments demonstrated significantly boosted performance in both the classic and the few-shot learning settings. 
\clearpage
\section{Limitations}
There are a number of areas that would warranty extensions or future work. First, a systematic study on the exact keywords and strategies of collecting such a data that would optimize the model outcome is yet to be studied. The methods proposed in this paper is merely a starting point for methods along this direction. Secondly, architecture changes that specifically targets the proposed methods of massive and noisy data collecting remains an open research question. One observation we had when examining the data is that many of them contains empty forms while others have filled in content. Models that can explicitly take advantage of both formats should further boost the performance of the model.

%% file: sections/appendix.tex
\section{Details on \modelname{} Models}
\label{app:model}
\input{sections/model_architectures}

\section{Additional Related Works}
\subsection{Datasets}
\label{app:related_data}

\paragraph{Smaller document datasets}
The Form Understanding in Noisy Scanned Documents (
FUNSD dataset~\cite{jaume2019funsd}, while being the most popular, only contains 199 document pages with three types of entities.
The
Consolidated Receipt Dataset for Post-OCR Parsing (
CORD~\cite{park2019cord} dataset comes at a larger scale with 1K document pages and 30 entity types.
Other datasets, such as 
the Scanned Receipts OCR and key Information Extraction (
SROIE~\cite{huang2019icdar2019}, Kleister~\cite{stanislawek2021kleister} NDA and Charity,
DeepForm~\cite{borchmann2021due},
VRDU~\cite{wang2022benchmark} Ad-buy and Registration, have been introduced since then, at the scale of a few thousand documents. 
Among them, DocVQA~\cite{mathew2021docvqa} contains 12.8K documents with question-answer annotations.

\paragraph{Larger document datasets}
IIT-CDIP~\cite{lewis2006building} consists of 11M unlabeled documents with more than 39M pages.
PDF files from Common Crawl (CC-PDF) and UCSF Industry Documents Library (UCSF-IDL) have also been used for pretraining~\cite{powalski2021going}, with a total of less than 1M documents.
RVL-CDIP, a subset of IIT-CDIP, contains 400K documents categorized into 16 classes for the document classification task.
PubLayNet~\cite{zhong2019publaynet} is at a similar scale but for the layout detection task with bounding box and segmentation annotations.

\subsection{Document Understanding Models}
\label{app:related_model}
Document understanding models have emerged since LayoutLM~\cite{xu2020layoutlm}, which extends BERT~\cite{devlin2019bert} with spatial and visual information.
Various models use different initialization weights, model scales, and pretraining data configurations. 
Table \ref{tab:sota} provides a detailed comparison of existing models.

\paragraph{Text Modality.}
Document models are usually built upon a pretrained language model.
As shown by LayoutLM~\cite{xu2020layoutlm}, language initialization significantly impacts the final model performance.
Many works have been built upon the standard BERT language model, such as LayoutLM~\cite{xu2020layoutlm}, BROS~\cite{hong2022bros}, SelfDoc~\cite{li2021selfdoc}, and UDoc~\cite{gu2021unidoc}.
LayoutLMv2~\cite{xu2021layoutlmv2} is initialized from the UniLM~\cite{dong2019unified}.
TILT~\cite{powalski2021going} extends T5~\cite{raffel2020exploring} for document analysis.
DocFormer~\cite{appalaraju2021docformer} directly initializes from a pretrained LayoutLM.
The recent LiLT~\cite{wang2022lilt} and LayoutLMv3~\cite{huang2022layoutlmv3} models are initialized from RoBERTa~\cite{liu2019roberta} to provide a stronger language prior.
In our experiments, we adopt the vanilla BERT-base model for fair comparisons without the benefit of a stronger language model.

\paragraph{Visual Modality.}
Existing document models rely on pretrained image models to utilize the document images.
LayoutLM~\cite{xu2020layoutlm} adopts a pretrained ResNet-101~\cite{he2016deep} as the visual feature encoder only during finetuning.
LayoutLMv2~\cite{xu2021layoutlmv2} further utilizes a ResNeXt-101~\cite{xie2017aggregated} at both pretraining and finetuning with encoded patch features as visual tokens.
In addition, SelfDoc~\cite{li2021selfdoc}, UDoc~\cite{gu2021unidoc}, TILT~\cite{powalski2021going}, and DocFormer~\cite{appalaraju2021docformer} also adopt a pretrained ResNet~\cite{he2016deep} as the visual feature encoder.
LayoutLMv3~\cite{huang2022layoutlmv3} distills a pretrained document image dVAE~\cite{ramesh2021zero} from DiT~\cite{li2022dit} to learn the visual modality during pretraining.
In contrast, we do not use pretrained image models but learn a joint vision-language representation by aligning both modalities at the token level.


\section{Detailed Experimental Setups and Analysis}
\subsection{Classic VDER Setting}
\label{app:classic}
\paragraph{Task setup.}
FUNSD contains 199 documents with 149 for training and 49 for evaluation. It is labeled with 3 entity types, i.e., header, question, and answer.
CORD contains 1000 documents with 800 for training, 100 for validation, and 100 for testing.
It is labeled with 30 entity types for receipts, such as menu name, price, \etc.
RVL-CDIP contains 400K documents in 16 classes, with 320K for training, 40K for validation, and 40K for testing.



\paragraph{Error analysis.}
\input{tables/funsd.tex}
\begin{figure*}[tp]
    \centering
    \includegraphics[width=\linewidth, clip]{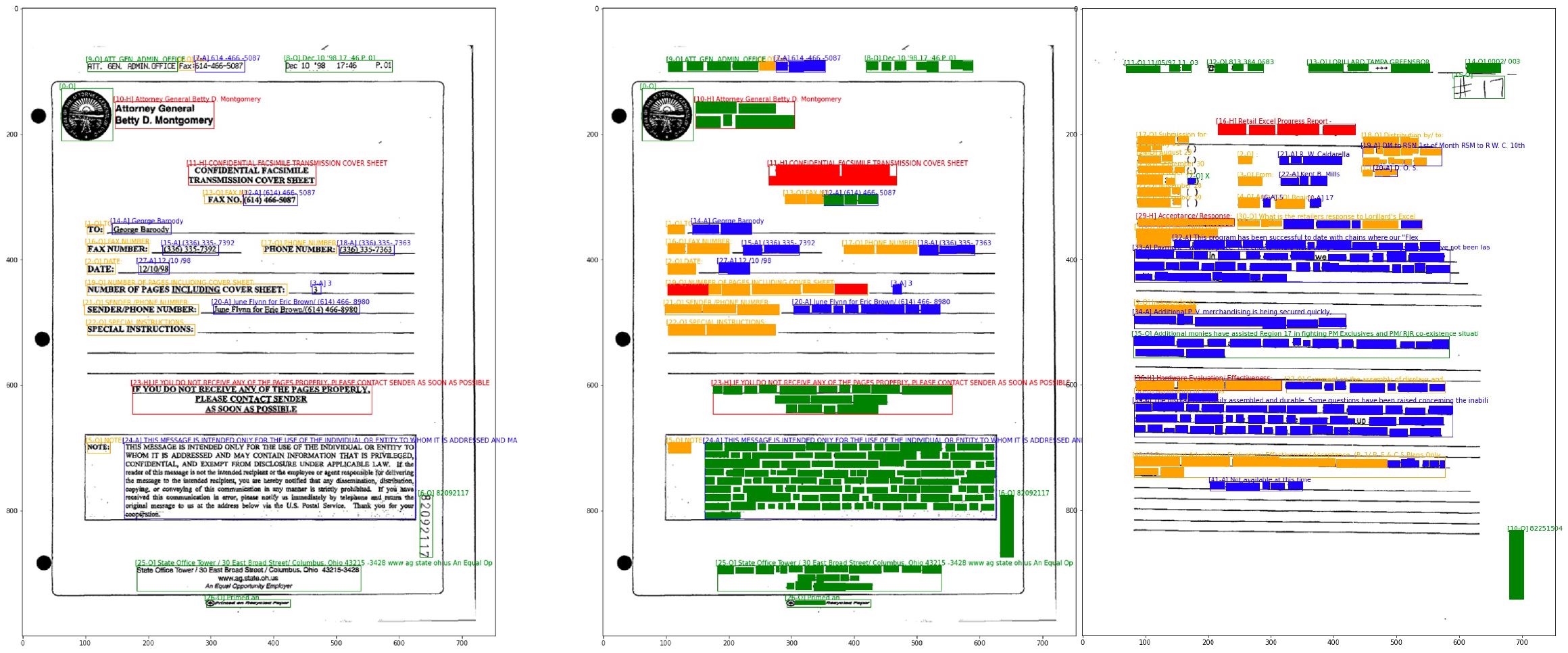}
    \caption{Visualization of annotation (left) and prediction examples (middle and right) from the FUNSD validation set. Zoom in for details.}
    \label{fig:sample}
    \vspace{-1em}
\end{figure*}
Table \ref{tab:funsd} lists the detailed metrics on the FUNSD entity extraction task.
Among the three labeled entity types, \textit{header} has the poorest performance and the lowest number of examples.
The other two types have much better performance with F1 86.59 for \textit{question} and F1 84.91 for \textit{answer}.
Fig. \ref{fig:sample} visualizes a few examples with annotations and predictions from our \modelname{}.
As we can see in the annotation, the reading order is often weird and does not follow human conventions.
However, the 2D positional embedding and spatial-aware attention can correctly handle them regardlessly.
In the prediction samples, we observe that the predictions for \textit{question} and \textit{answer} fields are mostly correct, while a few errors are made for \textit{header} due to ambiguity.

\subsection{Few-shot VDER Setting}
\label{app:few-shot}
\paragraph{N-way K-shot meta-learning formulation.}
In our setting, we define a $N$-way $K$-shot problem to be one such that there are $N$ novel classes that appear no more than $K$ times in the training set. We then divide a dataset into several sub-groups with each of them satisfying the $N$-way $K$-shot definition. One unique characteristic on the VDER dataset is that documents usually contain multiple entities, with many of the entities occur more than once in a single document, we make the requirements on the number of occurrence $K$ to be a soft one so that it would be realistic to generate such a dataset splitting. The few-shot learning problem will natually fit into a meta-learning scenario, meta-train and meta-test both contain a set of tasks satisfying N-way K-shot setting.

We sample datasets to achieve n-way, k-shot settings, which means that our training data contains n entities, each with at least k occurrences. The count of classes in testing is fixed at 5. For hyper-parameters, we follow most of the settings for classic VDER experiments. We fine tune with a learning rate of 0.02.



\section{\dataname{} Ontology}
\label{app:ontology}

\begin{figure*}
  \includegraphics[width=0.95\textwidth,trim={0 0.1cm 0 0.2cm},clip]{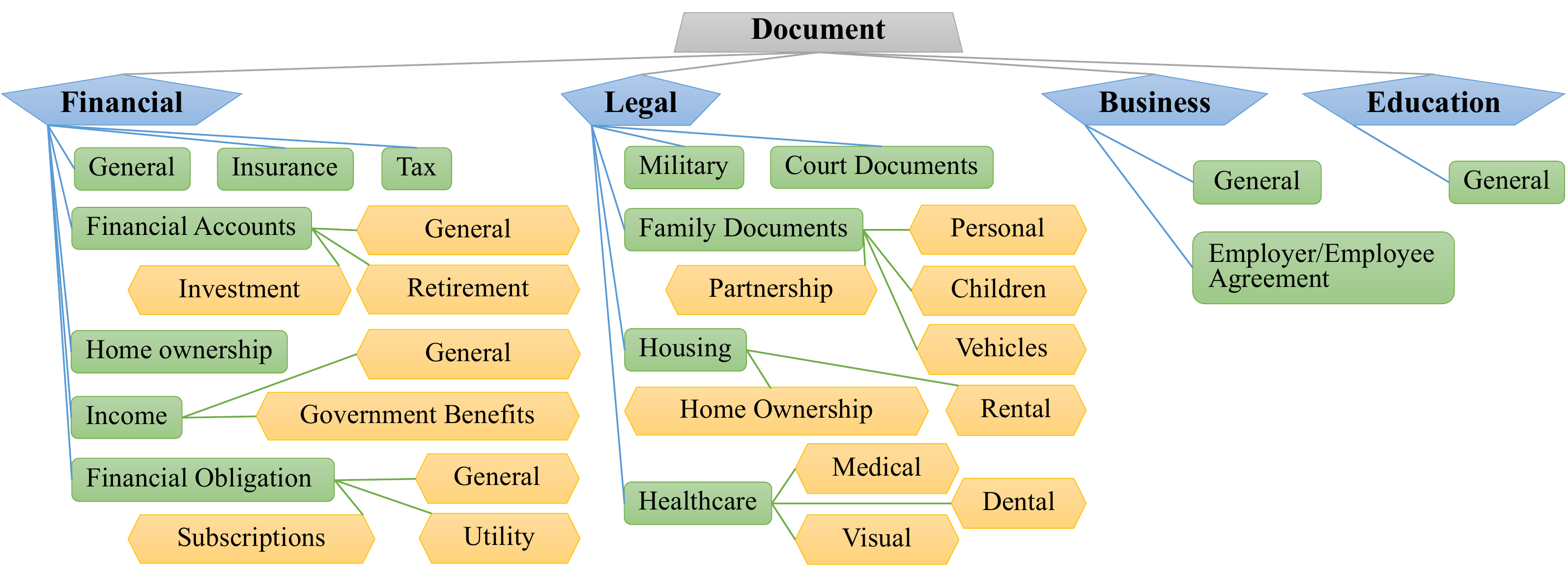}
  \vspace{-2mm}
  \caption{Document ontology tree stub, based on which the proposed \dataname{} datasets are collected. We create a document ontology with about 400 search keywords hierarchically connected by three intermediate layers. 
  }
  \label{fig:teaser}
  \vspace{-4mm}
\end{figure*}

Fig. \ref{fig:teaser} illustrates the document ontology tree stub used for the construction of \dataname{}.
Below we list all of the search keywords organized into four groups.

\input{sections/antology}

%% file: sections/model_architectures.tex
In this section, we detail our \modelname{} model architecture and setups for pretraining and fine-tuning for \taskname{}.

\subsection{Multimodal Tokenization}

Let $\rmD \in \sR^{H \times W \times 3}$ be a visually-rich document image with height $H$ and width $W$.
We obtain a sequence of characters by applying OCR on the document image.
The characters are accompanied by their bounding box coordinates.
Then we perform a multimodal tokenization process as follows.

With a pre-defined text tokenizer, we first tokenize the character sequence into a sequence of text tokens $\rvc$.
$\rvp$ represents the 1D position of the tokens ranging from $0$ to $|\rvc|-1$.
For each token $\ervc_i$, we obtain its bounding box $\ervb_i = (x_0, y_0, x_1, y_1)_i$ by taking the union of the bounding boxes of its characters.
We enlarge the bounding box by a context ratio $r$ on each side and obtain the corresponding visual image crop $\ervv_i$ for each token from $\rmD$.

\subsection{\modelname{} Architecture}

Fig. \ref{fig:pretrain_arch} illustrates the model architecture for our proposed \modelname{}.
\modelname{} is built upon BERT~\cite{devlin2019bert} and utilizes its tokenizer and pretrained weights.
The input for each token consists of a text embedding and a 1D position embedding for $\ervp$.

Following LayoutLM~\cite{xu2020layoutlm}, we add 2D position embeddings $x_0$, $y_0$, $x_1$, $y_1$, $w$, $h$, where $w = x_1 - x_0$ and $h = y_1 - y_0$.
These embeddings are used to represent the spatial location of each token.
All the embeddings mentioned above are obtained from trainable lookup tables.

Following LayoutLMv2~\cite{xu2021layoutlmv2}, \modelname{} adopts relative position-aware self-attention layers by adding biases to the attention scores according to relative 1D locations $\triangle \ervp$ and relative 2D locations $\triangle \frac{x_0 + x_1}{2}$, $\triangle \frac{y_0 + y_1}{2}$.

\paragraph{Image Crop Input}

\begin{figure}[tp]
\centering
\begin{subfigure}[b]{0.5\linewidth}
    \centering
    \includegraphics[width=\linewidth, clip]{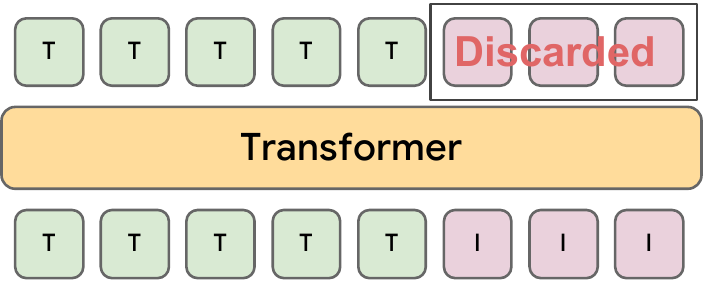}
\end{subfigure}
\hfill
\begin{subfigure}[b]{0.3\linewidth}
    \centering
    \includegraphics[width=\linewidth, clip]{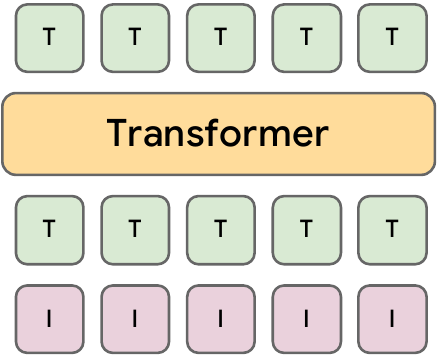}
\end{subfigure}
\caption{Unaligned (left) vs. Aligned (right) visual features. The unaligned visual features result in a longer sequence but are usually discarded in downstream tasks. T: Text, I: Image.}
\label{fig:align}
\end{figure}

To model visual information, we add a crop embedding by linearly projecting the flattened pixels in the image crop, following ViT~\cite{dosovitskiy2020image}.
Different from prior works using either uniform patches~\cite{huang2022layoutlmv3}, regional features~\cite{li2021selfdoc,gu2021unidoc}, or global features~\cite{appalaraju2021docformer}, our multimodal tokenization and linear embedding of image crops has the following advantages: 
\begin{itemize}[nosep, leftmargin=*]
    \item It eliminates the separate preprocessing for the visual modality, such as feature extraction with a pretrained CNN~\cite{xu2021layoutlmv2} or manually defined patches~\cite{huang2022layoutlmv3}.
    \item It obtains an aligned partition of the visual information with the text tokens, encouraging better cross-modal interaction.
    \item It eliminates the need for separate visual tokens as in \cite{xu2021layoutlmv2,huang2022layoutlmv3}, resulting in a shorter token sequence and better efficiency, as shown in Fig. \ref{fig:align}.
    \item It provides a unified joint representation for text and visual modalities in document modeling with semantic-level granularity.
\end{itemize}

\subsection{Pretraining}
During pretraining, we adopt the following objectives on a \modelname{} parameterized by $\theta$.
For each objective, we use a separate head upon the last attention layer.
Let $\rho$ denote the always available input embeddings, including the 1D and 2D positions.

\paragraph{Multimodal Masked Language Modeling (MMLM)}
We randomly select 15\%~\cite{devlin2019bert} of the tokens, denoted as $\gM$, to mask and predict the language modality.
In the masked language input $\overline{\rvc}$, 80\% of the masked tokens are replaced with a special \mask{} token, while another 10\% are replaced with a random token and the remaining 10\% are kept as is.
In the masked crop input $\overline{\rvp}$, crops for all masked tokens are replaced with an empty image.
The language prediction is formulated as a multi-class classification problem with the cross-entropy loss as
\begin{equation}
  \Ls_\mathit{MMLM} = \mathop{\E}\limits_{\rmD} \Big[ \sum_{\ervc_i \in \gM} -\log p_\theta (\ervc_i \mid [\overline{\rvc}, \overline{\rvv}, \rho]) \Big]
\end{equation}

\begin{figure*}
    \centering
    \includegraphics[width=\linewidth,trim={0 0 0.8cm 0},clip]{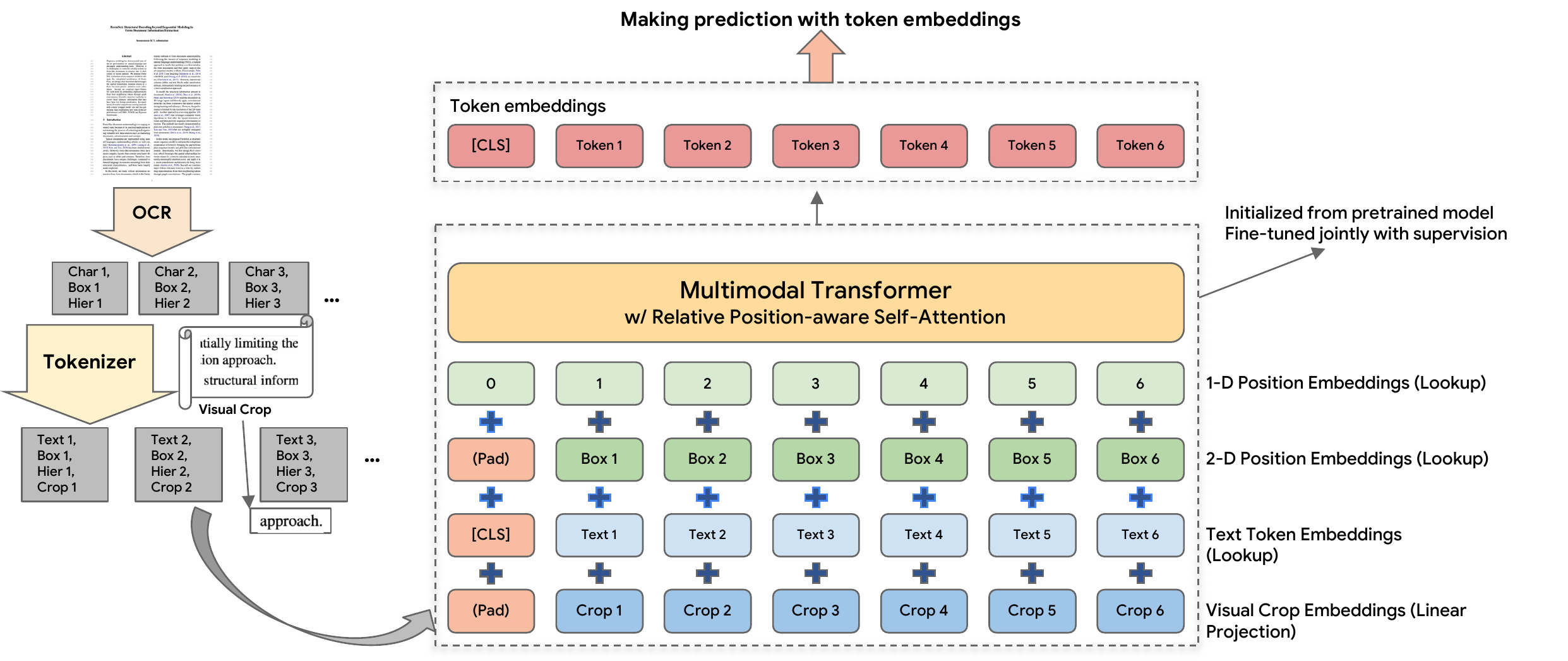}
    \caption{\modelname{} finetuning model architecture.}
    \label{fig:finetune_arch}
\end{figure*}

\paragraph{Masked Crop Modeling (MCM)}
We also predict the visual modality by reconstructing the image crops for the masked tokens in MMLM, in a way similar to MAE~\cite{he2022masked}.
It is formulated as a regression problem with a linear layer over flattened pixels.
The MCM loss is defined as 
\begin{equation}
    \Ls_\mathit{MCM} = \mathop{\E}\limits_{\rmD} \Big[ \sum_{\ervc_i \in \gM} \left\|\hat{\ervv}_i - \ervv_i\right\|_2^2 \Big]
\end{equation}
where $\hat{\rvv} = f_\theta(\overline{\rvc}, \overline{\rvv}, \rho])$.

\paragraph{Token Tagging (TT)}
We add an extra pretraining task by predicting the tags $\rvt$ for each token in an unmasked sequence.
The tags are extracted from an external text tagger as described in Sec. \ref{sec:ocr}.
Since each token may have multiple tags, it is formulated as a multi-label classification problem with the binary cross-entropy loss as 
\begin{equation}
\begin{aligned}
  \Ls_\mathit{TT} &= \mathop{\E}\limits_{\rmD} \Big[ \sum_{i, k} -\ervt_{i, k} \log p_\theta (\ervt_{i, k} \mid [\rvc, \rvv, \rho]) \\
  &-(1 - \ervt_{i, k}) \log (1 -  p_\theta (\ervt_{i, k} \mid [\rvc, \rvv, \rho])))  \Big]
\end{aligned}
\end{equation}
where $k=1,2,\cdots,K$ refers to the $K$ types of tags.

\paragraph{Pretraining Loss}
The overall pretraining objective is given as 
\begin{equation}
    \Ls_\mathit{pretrain} = \Ls_\mathit{MMLM} + \alpha \Ls_\mathit{MCM} + \beta \Ls_\mathit{TT}
\end{equation}
where $\alpha, \beta$ are the corresponding loss weights.

\subsection{Finetuning}

Fig. \ref{fig:finetune_arch} illustrates the pipeline for the finetuning of \modelname{}.
During finetuning, no tokens are masked.
In this paper, we adopt the following two tasks in finetuning.

\paragraph{Entity Extraction}
Entity extraction is formulated as a sequence tagging problem.
The ground-truth entity spans are converted into a sequence of BIO tags $\rve$ over all tokens.
The BIO tagging is formulated as follows:
$\rve$ is initialized with all $\gO$ tags which indicates ``Other" refering to background tokens.
For each entity span with type $\gT$, start position $i$ and end position $j$ (both inclusive), we assign 
\begin{align}
    \erve_i &= \gT_\text{Begin} \\
    \erve_{i+1} = ... = \erve_{j} &= \gT_\text{Intermediate}
\end{align}
The prediction of BIO tags is modeled as a multi-class classification problem with the objective as
\begin{equation}
  \Ls_\mathit{EE} = \mathop{\E}\limits_{\rmD} \Big[ \sum_{i} -\log p_\theta (\erve_i \mid [\rvc, \rvv, \rho]) \Big]
\end{equation}

\paragraph{Document Classification}
We use the embedding of the starting \cls token for document classification.
The logits are predicted with an MLP head on top of the \cls embedding.
Let $l$ be the correct class, the objective is
\begin{equation}
  \Ls_\mathit{DC} = \mathop{\E}\limits_{\rmD} \Big[ -\log p_\theta (l \mid [\rvc, \rvv, \rho]) \Big]
\end{equation}

%% file: tables/funsd.tex
\begin{table}[tp]
\begin{adjustbox}{max width=\linewidth}
\begin{tabular}{ccccc}
\hline
         & Precision & Recall & F1-score & Support \\ \hline
Question & 84.84    & 88.41 & 86.59   & 1070    \\
Header   & 57.26    & 56.30 & 56.78   & 119     \\
Answer   & 82.67    & 87.27 & 84.91   & 809     \\ \hline
Average  & 82.41    & 86.04 & \textbf{84.18}   & 1998    \\ \hline
\end{tabular}
\end{adjustbox}
\caption{Detailed metrics on the FUNSD entity extraction task.}
\label{tab:funsd}
\end{table}

%% file: sections/antology.tex
\subsection{Financial Documents}

\begin{itemize}[nosep]
\item accounts receivable aging report
\item bill of exchange pdf
\item invoice
\item receipt
\item loan estimate
\item loan application form
\item credit report pdf
\item employee insurance enrollment form
\item property insurance declaration page
\item renters insurance addendum
\item auto insurance card
\item dental insurance card
\item dental insurance verification form
\item vision insurance card
\item medical insurance card
\item liability insurance certificate
\item insurance cancellation letter
\item life insurance application form
\item flood elevation certificate
\item flood insurance application form
\item hazard insurance application form
\item tax return form
\item form 1040 schedule C
\item form 1040 schedule E
\item form 1040 schedule D
\item form 1040 schedule B
\item form 1040 nr
\item form 1040 sr
\item form 4506T EZ
\item form 4506T
\item form 4506 C
\item transfer of residence form 1076
\item property tax bill
\item W2
\item W4
\item 1099 B
\item 1099-MISC
\item 1099-NEC
\item 1099 DIV
\item 1099 G PDF
\item 1099 R
\item 1099 INT
\item SSA 1099 form
\item 1120 form
\item 1120S Form
\item form 1065
\item W7 form
\item W8BEN form
\item W9 form
\item SS4 form
\item form 940 pdf
\item form 5498
\item ucc 1 form
\item bank statement
\item personal check
\item check deposit slip pdf
\item credit union statement pdf
\item credit card authorization form
\item credit card
\item debit card
\item credit card statement
\item TSP election form
\item 401k enrollment form
\item IRA distribution request form
\item stock certificate
\item stock purchase agreement
\item bond certificate
\item bond purchase agreement
\item mutual fund consolidated account statement
\item HSA enrollment form
\item FSA enrollment form
\item verification of employment pdf
\item wage paystub
\item income verification letter
\item music recording contract
\item food stamp application form
\item us treasury check
\item child welfare services application form
\item medicaid card
\item medicaid application form
\item club application
\item membership renewal letter pdf
\item mortgage statement
\item rent invoice
\item electric bill
\item pg\&e care fera application pdf
\item gas bill
\item water bill
\item waste management invoice
\item spectrum internet bill pdf
\item phone bill pdf
\item car payment agreement
\item student loan payment agreement
\item child support agreement
\item child support receipt
\item elder care facility agreement
\item debt paymen tletter
\item demand for payment letter
\item magazine subscription form
\item streaming service agreement
\item gym waiver form
\item gym membership cancellation letter
\item gym membership card
\item massage therapy waiver
\item HOA agreement
\item HOA dues letter
\item urla form 1003
\item home appraisal report
\item security instrument
\item ucdp summary report
\item audit findings report pdf
\item sales contract
\item purchase agreement
\item title commitment pdf
\item earnest money deposit pdf
\item patriot act disclosure
\item owner occupancy affidavit form
\item compliance agreement
\item name affadavit
\item notice of right to reclaim abandoned property
\item VBA 26-0551 debt questionnaire pdf
\item VBA 26-8923 form pdf
\item USDA-AD 3030
\item loan application pdf
\item homeowner insurance declaration page
\item 1040
\item wage and tax statement
\item employee's withholding certificate
\item miscellaneous income form
\item nonemployee compensation form
\item dividends and distributions form
\item certain government payments
\item distributions from pensions
\item social security benefits form
\item form 1005
\item stimulus check pdf
\item waste management bill
\item comcast internet bill pdf
\item car loan payment agreement
\item gym release form
\item one and the same person affadavit
\item xfinity internet bill pdf
\item car payment contract
\end{itemize}

\subsection{Legal Documents}

\begin{itemize}[nosep]
\item birth certificate
\item social security card
\item social security form
\item ssa 89 form
\item social security change in information form
\item passport book
\item passport card
\item new passport application
\item passport renewal application
\item green card
\item green card application form
\item naturalization certificate
\item N-400 form pdf
\item living will sample
\item living will form
\item living will declaration
\item voter identification card
\item disability card
\item death certificate
\item death certificate application
\item name change form
\item state issued identification card
\item prenup form
\item postnuptial agreement
\item marriage license
\item marriage certificate
\item application for marriage license
\item family court cover sheet
\item complaint for divorce no children
\item complaint for divorce with children
\item divorce summons
\item divorce certificate
\item domestic partnership application form
\item domestic partnership certificate
\item domestic partnership termination form
\item separation agreement
\item pet custody agreement form 
\item pet ownership transfer form
\item child adoption certificate
\item child power of attorney
\item child visitation form
\item daycare contract
\item child custody agreement
\item child support modification form
\item free minor travel consent form
\item child identity card
\item DNA paternity test order form
\item petition for declaration of emancipation of minor
\item vehicle registration card
\item vehicle registration form
\item vehicle registration renewal notice
\item vehicle certificate of title
\item motor vehicle transfer form
\item driver's license
\item application for driver's license
\item truck driver application
\item learner's permit card
\item pilot's license card
\item vehicle leasing agreement
\item motor vehicle power of attorney
\item mortgage interest credit form
\item mortgage application form
\item mortgage verification form
\item mortgage loan modification form
\item real estate deed of trust
\item mortgage deed
\item warranty deed
\item quitclaim deed
\item usps mail forwarding form PDF
\item property power of attorney
\item notice of intent to foreclose
\item closing disclosure
\item HUD 92541 form
\item HUD 54114 form
\item HUD 92561 form
\item FHA loan underwriting and transmittal summary
\item form HUD92051
\item form HUD 92900-A
\item form HUD 92544
\item form HUD 92900 B important notice to housebuyers
\item form HUD 92900 WS mortgage credit analysis worksheet
\item Form HUD 92800 Conditional Commitment
\item SFHDF
\item lease agreement
\item lease application
\item notice to enter
\item notice of intent to vacate premises
\item notice of lease violation
\item pay rent or quit
\item lease offer letter
\item roommate agreement
\item eviction notice form
\item lease termination letter
\item lease renewal agreement
\item pet addendum
\item notice of rent increase
\item sublease agreement
\item record of immunization
\item allergy record sheet
\item allergy immunotherapy record
\item medication log
\item prescription sheet
\item disability documentation
\item advance directive form
\item DNA test request form
\item medical power of attorney
\item health care proxy form
\item revocation of power of attorney
\item dnr form
\item hipaa release form
\item hipaa complaint form
\item health history form
\item birth plan form
\item new patient form
\item child medical consent
\item grandparent medical consent for minor
\item medical treatment authorization form
\item dental policy and procedure document
\item endodontic treatment consent form
\item denture treatment consent form
\item dental patient referral form
\item patient dismissal letter
\item dental record release form
\item oral surgery postop instructions
\item refusal of dental treatment form
\item tooth extraction consent form
\item corrective lens prescription pdf
\item military id card
\item dd214
\item honorable discharge certificate
\item supreme court distribution schedule pdf
\item case docket pdf
\item jury summons pdf
\item jury duty excuse letter
\item supplemental juror information pdf
\item attorney termination letter
\item certificate of good standing
\item attorney oath of admission pdf
\item substitution of attorney
\item notice of appearance of counsel
\item bankruptcy declaration form
\item notice of lawsuit letter
\item court summons pdf
\item arrest warrant pdf
\item promissory note
\item tolling agreement pdf
\item notary acknowledgement form
\item cease and desist letter
\item condominium rider pdf
\item adjustable rate rider pdf
\item family rider form 1-4 pdf
\item balloon rider form pdf
\item second home rider pdf
\item revocable trust rider form pdf
\item pud rider pdf
\item birth certificate form
\item ssn card
\item application for a social security card
\item application for naturalization pdf
\item legal name change form
\item state issued ID
\item prenup sample
\item postnuptial agreement sample
\item declaration of domestic partnership
\item marriage separation agreement
\item minor power of attorney
\item request for child custody form
\item child care contract
\item motion to adjust child support
\item child travel consent form
\item vehicle registration application
\item vehicle certificate of title
\item driver's license application
\item mortgage loan application form
\item real estate power of attorney
\item foreclosure letter notice
\item rental agreement
\item rental application
\item landlord notice to enter
\item intent to vacate rental
\item notice to pay rent or quit
\item roommate contract
\item eviction notice pdf
\item early lease termination letter
\item pet addendum to lease agreement
\item rent increase letter
\item vaccine record form
\item prescription sample
\item healthcare directive form
\item do not resuscitate form
\item medical records release form
\item medical history form
\item new patient registration form
\item child medical release form
\item root canal consent form
\item eyeglasses prescription pdf
\item military discharge form
\item notice of intent to sue
\item ssn application
\item name change form example
\item prenuptial agreement sample
\item marital separation form
\item motion to modify child support
\item tenant application
\item notice to enter premises
\item notice to vacate
\item notice to quit
\item notice of lease termination
\item doctor prescription
\item patient history form
\item patient intake form
\item consent to treat minor
\item form petition for name change
\item medical intake form
\end{itemize}

\subsection{Business Documents}

\begin{itemize}[nosep]
\item articles of incorporation
\item corporate bylaws
\item operating agreement
\item shareholder agreement
\item memorandum of understanding
\item expense report
\item purchase of business agreement
\item purchase order
\item invoice pdf
\item late payment reminder letter
\item arbitration agreement pdf
\item business contract
\item payment agreement document
\item end user license agreement
\item licensing agreement pdf
\item job application form
\item employment offer letter
\item employment rejection letter
\item employment agreement
\item resume
\item employment resignation letter
\item notice of contract termination
\item notice of employmen termination
\item nda pdf
\item non compete agreement
\item leave of absence request
\item employment evaluation form
\item overdue payment reminder letter
\item job application pdf
\item job offer letter
\item job rejection letter
\item employment contract
\item contract termination letter
\item non disclosure agreement
\end{itemize}

\subsection{Education Documents}

\begin{itemize}[nosep]
\item research papers pdf
\item certificate of enrollment
\item high school transcript
\item high school diploma
\item college diploma
\item college transcript
\end{itemize}